\documentclass[11pt,a4paper]{article}
\usepackage[hyperref]{acl2021}
\usepackage{times}
\usepackage{latexsym}
\usepackage{multirow}
\usepackage{amsmath}
\usepackage{amssymb}
\usepackage{bbm}
\usepackage{graphicx}
\usepackage{algorithm}
\usepackage{algorithmicx}
\usepackage{algpseudocode}
\usepackage{paralist}
\usepackage{bm}
\usepackage{enumitem}
\usepackage{subfigure}
\usepackage{pifont}

\newcommand\mydots{\makebox[1em][c]{.\hfil.\hfil.}}

\usepackage{microtype}

\aclfinalcopy 
\def\aclpaperid{} 

\newcommand\model{\texttt{MTM}}

\title{Article Reranking by Memory-Enhanced Key Sentence Matching \\ for Detecting Previously Fact-Checked Claims}

\author{Qiang Sheng$^{1,2}$,  Juan Cao$^{1,2}$, Xueyao Zhang$^{1,2}$, Xirong Li$^{3}$, Lei Zhong$^{1,2}$  
\\
 \textsuperscript{1} Key Laboratory of Intelligent
Information Processing, \\
 Institute of Computing Technology, Chinese Academy of Sciences\\
  \textsuperscript{2}University of Chinese Academy of Sciences\\
  \textsuperscript{3}Key Lab of Data Engineering and Knowledge Engineering, Renmin University of China\\
  \texttt{\{shengqiang18z,caojuan,zhangxueyao19s,zhonglei18s\}@ict.ac.cn}\\ \texttt{xirong@ruc.edu.cn}
}

\date{}

\begin{document}
\maketitle

\begin{abstract}
False claims that have been previously fact-checked can still spread on social media. To mitigate their continual spread, detecting previously fact-checked claims is indispensable.
Given a claim, existing works retrieve fact-checking articles (FC-articles) for detection and focus on reranking candidate articles in the typical two-stage retrieval framework.
However, their performance may be limited as they ignore the following characteristics of FC-articles: (1) claims are often quoted to describe the checked events, providing lexical information besides semantics; and (2) sentence templates to introduce or debunk claims are common across articles, providing pattern information.
In this paper, we propose a novel reranker, \model~(\underline{M}emory-enhanced \underline{T}ransformers for \underline{M}atching), to rank FC-articles using key sentences selected using event (lexical and semantic) and pattern information.
For event information, we propose to finetune the Transformer with regression of \texttt{ROUGE}. For pattern information, we generate pattern vectors as a memory bank to match with the parts containing patterns. 
By fusing event and pattern information, we select key sentences to represent an article and then predict if the article fact-checks the given claim using the claim, key sentences, and patterns.
Experiments on two real-world datasets show that \model~outperforms existing methods. Human evaluation proves that \model~can capture key sentences for explanations. The code and the dataset are at \url{https://github.com/ICTMCG/MTM}.
\end{abstract}

\begin{figure}[ht]
	\centering
	\includegraphics[width=\linewidth]{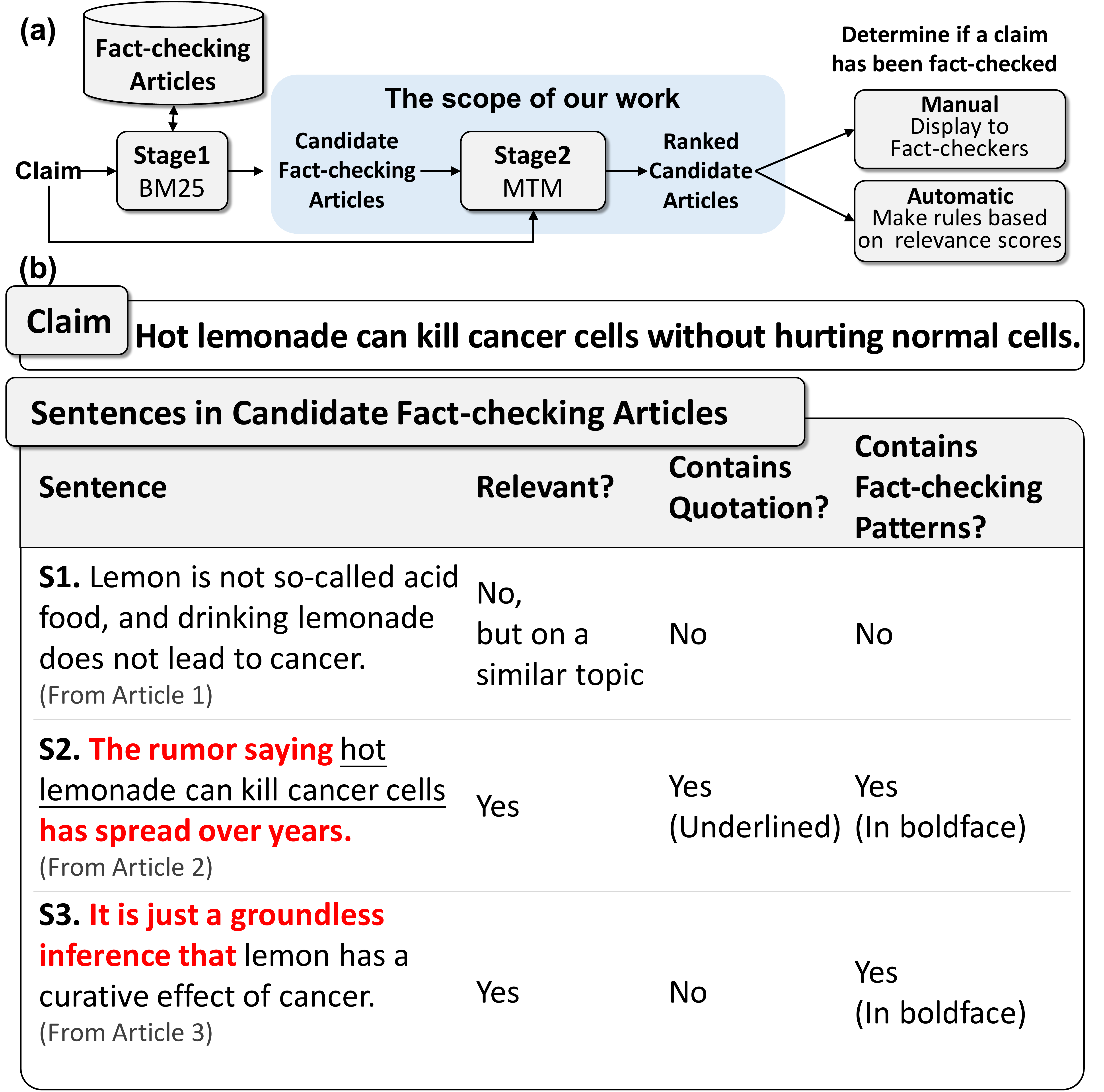}
	\caption{(a) Workflow of detecting a previously fact-checked claim. Our model \model~focuses on the second stage, i.e., reranking the candidates. (b) A claim and sentences in the candidate fact-checking articles (translated from Chinese). S1 is on a similar topic but actually irrelevant, while S2 and S3 which contain quotation or fact-checking patterns are relevant.}
	\label{fig1}
\end{figure}

\section{Introduction}\label{sec:intro}
Social media posts with false claims have led to real-world threats on many aspects such as politics~\citep{pizzagate}, social order~\citep{salt}, and personal health~\citep{shuanghuanglian}.
To tackle this issue, over 300 fact-checking projects have been launched, such as Snopes\footnote{\url{https://www.snopes.com}} and Jiaozhen\footnote{\url{https://fact.qq.com/}}~\citep{duke-report}. Meanwhile, automatic systems have been developed for detecting suspicious claims on social media~\citep{newsverify, credeye}.
This is however not the end. A considerable amount of false claims continually spread, even though they are already proved false. 
According to a recent report~\citep{tencent-report}, around 12\% of false claims published on Chinese social media, are actually ``old'', as they have been debunked previously. Hence, detecting previously fact-checked claims is an important task. 

According to the seminal work by \citet{shaar2020}, the task is tackled by a two-stage information retrieval approach. Its typical workflow is illustrated in  Figure~\ref{fig1}(a).
Given a claim as a query, in the first stage a basic searcher (e.g., BM25 \citealp{bm25}) searches for candidate articles from a collection of fact-checking articles (FC-articles). In the second stage, a more powerful model (e.g.,  BERT,~\citealp{bert}) reranks the candidates to provide evidence for manual or automatic detection.
Existing works focus on the reranking stage: \citet{vo2020} model the interactions between a claim and the whole candidate articles, while \citet{shaar2020} extract several semantically similar sentences from FC-articles as a proxy.
Nevertheless, these methods treat FC-articles as \emph{general} documents and ignore characteristics of FC-articles. Figure~\ref{fig1}(b) shows three sentences from candidate articles for the given claim. Among them, S1 is more friendly to semantic matching than S2 and S3 because the whole S1 focuses on describing its topic and does not contain tokens irrelevant to the given claim, e.g., "has spread over years" in S2. Thus, a semantic-based model does not require to have strong filtering capability. If we use only general methods on this task, the relevant S2 and S3 may be neglected while irrelevant S1 is focused. To let the model focus on key sentences (i.e., sentences as a good proxy of article-level relevance) like S2 and S3, we need to consider two characteristics of FC-articles besides semantics: \textbf{C1}. Claims are often quoted to describe the checked events (e.g., the underlined text in S2); \textbf{C2}. Event-irrelevant patterns to introduce or debunk claims are common in FC-articles (e.g., bold texts in S2 and S3).

Based on the observations, we propose a novel reranker, \model~(\underline{M}emory-enhanced \underline{T}ransformers for \underline{M}atching). The reranker identifies key sentences per article using claim- and pattern-sentence relevance, and then integrates information from the claim, key sentences, and patterns for article-level relevance prediction. In particular, regarding \textbf{C1}, we propose \texttt{ROUGE}-guided Transformer (ROT) to score claim-sentence relevance literally and semantically. 

As for \textbf{C2}, we obtain the pattern vectors by clustering the difference of sentence and claim vectors for scoring pattern-sentence relevance and store them in the Pattern Memory Bank (PMB).
The joint use of ROT and PMB allows us to identify key sentences that reflect the two characteristics of FC-articles. 
Subsequently, fine-grained interactions among claims and key sentences are modeled by the multi-layer Transformer and aggregated with patterns to obtain an article-level feature representation. The article feature is fed into a Multi-layer Perceptron (MLP) to predict the claim-article relevance.

To validate the effectiveness of our method, we built the first Chinese dataset for this task with 11,934 claims collected from Chinese Weibo\footnote{\url{https://weibo.com}} and 27,505 fact-checking articles from multiple sources. 39,178 claim-article pairs are annotated as relevant. Experiments on the English dataset and the newly built Chinese dataset show that \model~outperforms existing methods. Further human evaluation and case studies prove that \model~ finds key sentences as explanations. Our main contributions are as follows:
\begin{compactitem}
	\item We propose a novel reranker \model~for fact-checked claim detection, which can better identify key sentences in fact-checking articles by exploiting their characteristics.
	\item We design \texttt{ROUGE}-guided Transformer to combine lexical and semantic information and propose a memory mechanism to capture and exploit common patterns in fact-checking articles.  
	\item Experiments on two real-world datasets show that \model~outperforms existing methods. Further human evaluation and case studies prove that our model finds key sentences as good explanations.
	\item We built the first Chinese dataset for fact-checked claim detection with fact-checking articles from diverse sources.
\end{compactitem}

\begin{figure*}[ht]
	\centering
	\includegraphics[width=\textwidth]{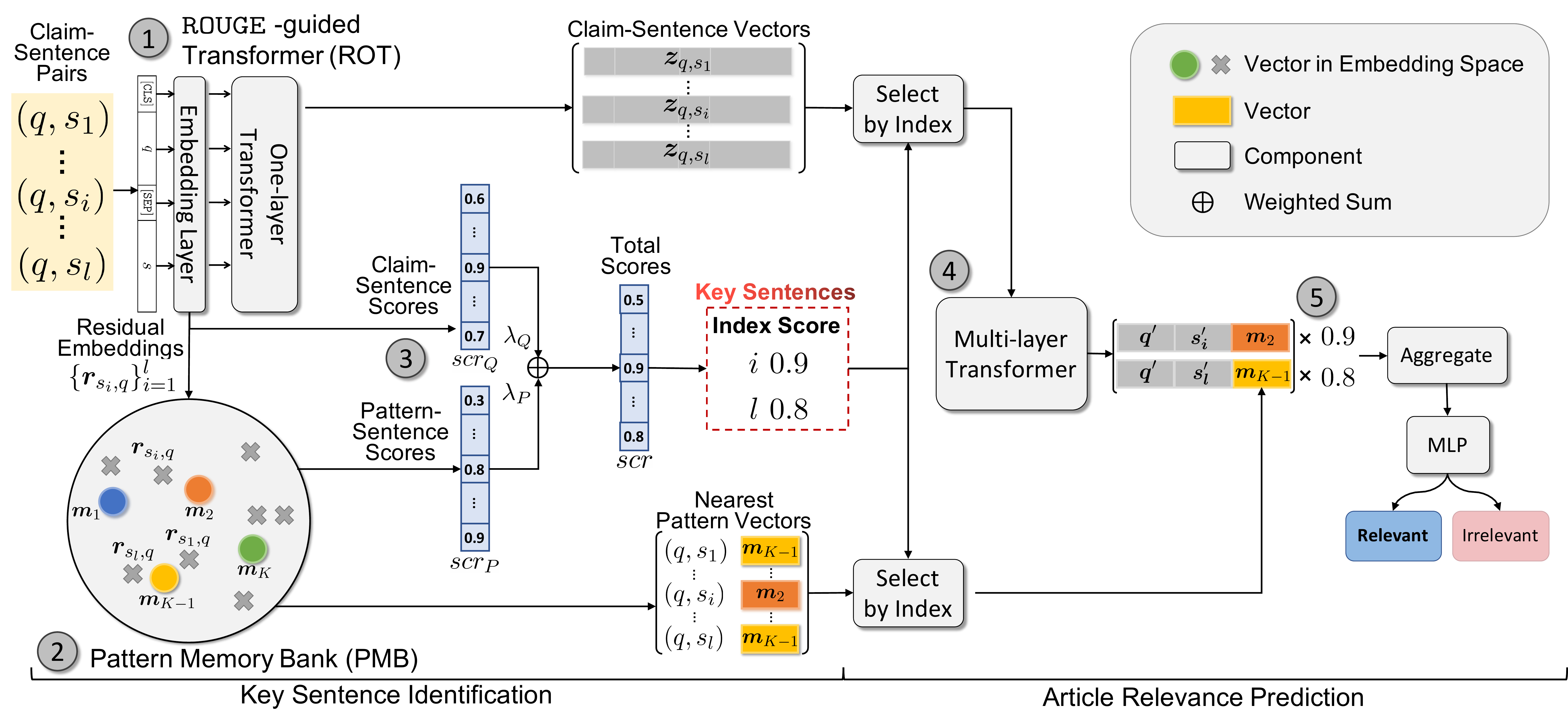}
	\caption{Architecture of \model. Given a claim $q$ and a candidate article $d$ with $l$ sentences, $s_1,\mydots,s_l$, \model~\ding{172} feeds $(q,s)$ pairs into \texttt{ROUGE}-guided Transformer (ROT) to obtain claim-sentence scores in both lexical and semantic aspects; \ding{173} matches residual embeddings $\bm{r}_{s,q}$ with vectors in Pattern Memory Bank (PMB) (here, only four are shown) to obtain pattern-sentence scores; \ding{174} identifies $k_2$ key sentences by combining the two scores (here, $k_2=2$, and $s_i$ and $s_l$ are selected); \ding{175} models interaction among $\bm{q^\prime},\bm{s^\prime},$ and the nearest memory vector $\bm{m}$ for each key sentence; and \ding{176} perform score-weighted aggregation and predict the claim-article relevance.}
	\label{fig:arch}
\end{figure*}

\section{Related Work}\label{sec:rw}
To defend against false information, researchers are mainly devoted to two threads: (1) \textbf{Automatic fact-checking} methods mainly retrieve relevant factual information from designated sources and judge the claim's veracity. \citet{fever} use Wikipedia as a fact tank and build a shared task for automatic fact-checking, while \citet{declare} and \citet{RDD} retrieve webpages as evidence and use their stances on claims for veracity prediction. (2) \textbf{Fake news detection} methods often use non-factual signals, such as styles~\citep{style-aaai,style-image}, emotions~\citep{ajao, dual-emotion}, source credibility~\citep{fang}, user response~\citep{defend} and diffusion network~\citep{early, alone}. However, these methods mainly aim at newly emerged claims and do not address those claims that have been fact-checked but continually spread. Our work is in a new thread, \textbf{detecting previously fact-checked claims}. \citet{vo2020} models interaction between claims and FC-articles by combining GloVe~\citep{glove} and ELMo embeddings~\citep{elmo}. \citet{shaar2020} train a RankSVM with scores from BM25 and Sentence-BERT for relevance prediction. These methods ignore the characteristics of FC-articles, which limits the ranking performance and explainability. 

\section{Proposed Method}\label{sec:method}

Given a claim $q$ and a candidate set of $k_1$ FC-articles $\mathcal{D}$ obtained by a standard full-text retrieval model (BM25), we aim to rerank FC-articles truly relevant w.r.t. $q$ at the top by modeling fine-grained relevance between $q$ and each article $d\in \mathcal{D}$. This is accomplished by Memory-enhanced Transformers for Matching (\model), which conceptually has two steps, (1) Key Sentence Identification and (2) Article Relevance Prediction, see Figure~\ref{fig:arch}.
For an article of $l$ sentences, let $\mathcal{S}=\{s_{1},\mydots,s_{l}\}$ be its sentence set.  
In Step (1), for each sentence, we derive claim-sentence relevance score from \texttt{ROUGE}-guided Transformer (ROT) and pattern-sentence relevance score from Pattern Memory Bank (PMB). The scores indicate how similar the sentence is to the claim and pattern vectors, i.e., how possible to be a key sentence. Top $k_2$ sentences are selected for more complicated interactions and aggregation with the claim and pattern vectors in Step (2). The aggregated vector is used for the final prediction. We detail the components and then summarize the training procedure below.
 
\subsection{Key Sentence Identification}
\subsubsection{\texttt{ROUGE}-guided Transformer (ROT)}

ROT (left top of Figure.~\ref{fig:arch}) is used to evaluate the relevance between $q$ and each sentence $s$ in $\{\mathcal{S}_i\}^{k_1}_{i=1}$, both lexically and semantically. 
Inspired by \citep{complementing}, we choose to ``inject'' the ability to consider lexical relevance into the semantic model. As the BERT is proved to capture and evaluate semantic relevance~\citep{BERTScore}, we use a one-layer Transformer initialized with the first block of pretrained BERT to obtain the initial semantic representation of $q$ and $s$:
\begin{equation}
	{\bm z}_{q,s} = \mathrm{Transformer}\left(\mathtt{[CLS]}\ q\ \mathtt{[SEP]}\ s \right)
\end{equation}
where $\mathtt{[CLS]}$ and $\mathtt{[SEP]}$ are preserved tokens and $\bm{z}_{q,s}$ is the output representation.

To force ROT to consider the lexical relevance, we finetune the pretrained Transformer with the guidance of \texttt{ROUGE}~\citep{rouge}, a widely-used metric to evaluate the lexical similarity of two segments in summarization and translation tasks. The intuition is that lexical relevance can be characterized by token overlapping, which ROUGE exactly measures. We minimize the mean square error between the prediction and the precision and recall of \texttt{ROUGE-2} between $q$ and $s$ ($\texttt{R}_{2} \in \mathbb{R}^2$) to optimize the ROT:
\begin{equation}
    \hat{\texttt{R}}(q,s)=\mathrm{MLP}\big(\bm{z}_{q,s}(\mathtt{[CLS]})\big)
\end{equation}
\begin{equation}
    \mathcal{L}_{R} = \Vert\hat{\texttt{R}}(q,s)-\texttt{R}_{2}(q,s)\Vert^2_2 + \lambda_R\Vert\Delta\theta\Vert^2_2
\end{equation}
where the first term is the regression loss and the second is to constraint the change of parameters as the ability to capture semantic relevance should be maintained. $\lambda_R$ is a control factor and $\Delta\theta$ represents the change of parameters.

\subsubsection{Pattern Memory Bank (PMB)}
The Pattern Memory Bank (PMB) is to generate, store, and update the vectors which represent the common patterns in FC-articles. The vectors in PMB will be used to evaluate pattern-sentence relevance (see~Section \ref{sec:selection}). Here we detail how to formulate, initialize, and update these patterns below.

\noindent\textbf{Formulation.} 
Intuitively, one can summarize the templates, like ``\mydots has been debunked by\mydots'', and explicitly do \emph{exact} matching, but the templates are costly to obtain and hard to integrate into neural models.
Instead, we \emph{implicitly} represent the common patterns using vectors derived from embeddings of our model, ROT. 
Inspired by~\citep{memory-bank}, we use a memory bank $\mathcal{M}$ to store $K$ common patterns (as vectors), i.e., $\mathcal{M}=\{\bm{m}_i\}^{K}_{i=1}$.

\noindent\textbf{Initialization.} We first represent each $q$ in the training set and $s$ in the corresponding articles by averaging its token embeddings (from the embedding layer of ROT).
Considering that a pattern vector should be \emph {event-irrelevant}, we heuristically remove the event-related part in $s$ as possible by calculating the residual embeddings $\bm{r}_{s,q}$, i.e., subtracting $\bm{q}$ from $\bm{s}$.
We rule out the residual embeddings that do not satisfy $t_{low}<\left\|\bm{r}_{s,q}\right\|_2<t_{high}$, because they are unlikely to contain good pattern information: $\left\|\bm{r}_{s,q}\right\|_2\leq t_{low}$ indicates $q$ and $s$ are highly similar and thus leave little pattern information, while $\left\|\bm{r}_{s,q}\right\|_2\geq t_{high}$ indicates $s$ may not align with $q$ in terms of the event, so the corresponding $r_{s,q}$ is of little sense.
Finally, we aggregate the valid residual embeddings into $K$ clusters using K-means and obtain the initial memory bank $\mathcal{M}$:
\begin{equation}
	\mathcal{M}=\mathrm{K}\mbox{-}\mathrm{means}\big(\{\bm{r}_{s,q}^{valid}\}\big) \!=\! \{\bm{m}_1,\mydots,\bm{m}_K\}
\end{equation}
where $\{\bm{r}_{s,q}^{valid}\}$ is the set of valid residual embeddings.

\noindent\textbf{Update.} As the initial $K$ vectors may not accurately represent common patterns, we update the memory bank according to the feedbacks of results during training: If the model predicts rightly, the key sentence, say $s$, should be used to update its nearest pattern vector $m$. To maintain stability, we use an epoch-wise update instead of an iteration-wise update.

\begin{figure}[t]
	\centering
	\includegraphics[width=0.43\textwidth]{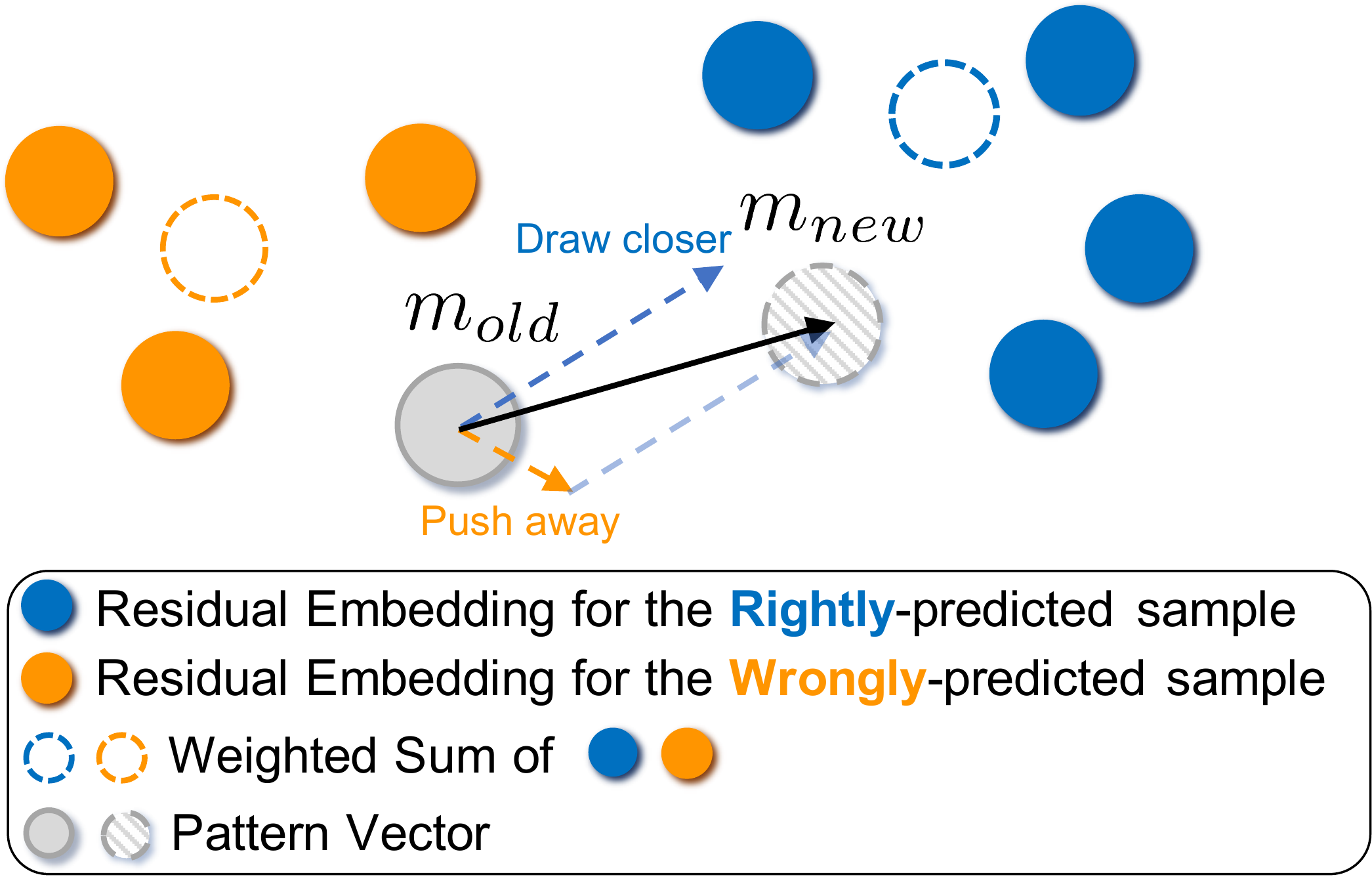}
	\caption{Illustration for Memory Vector Update.}
	\label{fig:mem_update}
\end{figure}

Take updating $\bm{m}$ as an example. After an epoch, we extract all $n$ key sentences whose nearest pattern vector is $\bm{m}$ and their $n$ corresponding claims, which is denoted as a tuple set $(\mathcal{S}, \mathcal{Q})^m$. Then $(\mathcal{S}, \mathcal{Q})^m$ is separated into two subsets, $\mathcal{R}^{m}$ and $\mathcal{W}^{m}$, which contain $n_r$ and $n_w$ sentence-claim tuples from the rightly and wrongly predicted samples, respectively. The core of our update mechanism (Figure~\ref{fig:mem_update}) is to draw $\bm{m}$ closer to the residual embeddings in $\mathcal{R}^{m}$ and push it away from those in $\mathcal{W}^{m}$. We denote the $i^{th}$ residual embedding from the two subsets as $\bm{r}_{\scriptscriptstyle \mathcal{R}_i^m}$ and $\bm{r}_{\scriptscriptstyle \mathcal{W}_i^m}$, respectively.

To determine the update direction, we calculate a weighted sum of residual embeddings according to the predicted matching scores.
For $(s,q)$, suppose \model~output $\hat{y}_{s,q} \in [0,1]$ as the predicted matching score of $q$ and $d$ (whose key sentence is $s$), the weight of $\bm{r}_{s,q}$ is $|\hat{y}_{s,q}-0.5|$ (denoted as $w_{s,q}$). Weighted residual embeddings are respectively summed and normalized as the components of the direction vector (Eq.~\eqref{eq:aggregated_res}):
\begin{equation}
\label{eq:aggregated_res}
	\bm{u}^{mr}\!= \!\bigg( \sum_{i=1}^{n_r}  w_{\scriptscriptstyle \mathcal{R}_i^m}\bm{r}_{\scriptscriptstyle \mathcal{R}_i^m} \bigg), 
	\bm{u}^{mw}\!=\!\bigg( \sum_{i=1}^{n_w}  w_{\scriptscriptstyle \mathcal{W}_i^m} \bm{r}_{\scriptscriptstyle \mathcal{W}_i^m}\bigg)
\end{equation}
where $\bm{u}^{mr}$ and $\bm{u}^{mw}$ are the aggregated residual embeddings. The direction is determined by Eq.~\eqref{eq:direction}:
\begin{equation}\label{eq:direction}
	\bm{u}^{m}=w_r\underbrace{(\bm{u}^{mr}-\bm{m})}_{\mbox{\small{draw closer}}}+ w_w\underbrace{(\bm{m}-\bm{u}^{mw})}_{\mbox{\small{push away}}}
\end{equation}
where $w_r$ and $w_w$ are the normalized sum of corresponding weights used in Eq.~\eqref{eq:aggregated_res} ($w_r+w_w=1$).
The pattern vector $m$ is updated with:
\begin{equation}
\label{eq:memory-update}
	\bm{m}_{new} = \bm{m}_{old} + \lambda_m \Vert\bm{m}_{old}\Vert _2 \frac{\bm{u}^{m}}{\Vert\bm{u}^{m}\Vert_2}
\end{equation}
where $\bm{m}_{old}$ and $\bm{m}_{new}$ are the memory vector $\bm{m}$ before and after updating; the constant $\lambda_m$ and $\left\|\bm{m}_{old}\right\|_2$ jointly control the step size. 

\subsubsection{Key Sentence Selection}\label{sec:selection}
Whether a sentence is selected as a key sentence is determined by combining claim- and pattern-sentence relevance scores. The former is calculated with the distance of $q$ and $s$ trained with ROT (Eq.~\eqref{eq:score_q}) and the latter uses the distance between the nearest pattern vector in PMB and the residual embedding (Eq.~\eqref{eq:score_p}). The scores are scaled to $[0,1]$. For each sentence $s$ in $d$, the relevance score with $q$ is calculated by Eq.~\eqref{eq:score_all}:
\begin{equation}
	scr_Q(q,s)=\mathrm{Scale}(\left\|\bm{r}_{s,q}\right\|_2)\\ \label{eq:score_q}
\end{equation}
\begin{equation}\label{eq:score_p}
	scr_P(q,s)=\mathrm{Scale}(\left\|\bm{m}_{u}-\bm{r}_{s,q}\right\|_2)
\end{equation}
\begin{equation}\label{eq:score_all}
	scr(q,s)=\lambda_{Q}scr_Q(q,s)+\lambda_{P}scr_P(q,s)
\end{equation}
where $\mathrm{Scale}(x)\!=\!1-\frac{x-min}{max-min}$ and $max$ and $min$ are the maximum and minimum distance of $s$ in $d$, respectively. $u=\arg \min_{i} \left\|\bm{m}_i-\bm{r}_{s,q}\right\|_2$, and $\lambda_{Q}$ and $\lambda_{P}$ are hyperparameters whose sum is $1$. 

Finally, sentences with top-$k_2$ scores, denoted as $\mathcal{K}\! =\!\{s^{key}_i(q,d)\}^{k_2}_{i=1}$, are selected as the \emph{key sentences} in $d$ for the claim $q$.

\subsection{Article Relevance Prediction (ARP)}\label{sec:reasoner}
\noindent\textbf{Sentence representation.} We model more complicated interactions between the claim and the key sentences by feeding each $z_{q,s^{key}}$ (derived from ROT) into a multi-layer Transformer ($\mathrm{MultiTransformer}$):
\begin{equation}
\label{eq:reasoner1}
	\bm{z}^\prime_{q,s^{key}}= \mathrm{MultiTransformer}(\bm{z}_{q,s^{key}})
\end{equation}
Following~\citep{sentence-bert}, we respectively compute the mean of all output token vectors of $q$ and $s$ in $z^\prime_{q,s^{key}}$ to obtain the fixed sized sentence vectors $\bm{q}^\prime \in \mathbb{R}^{dim}$ and $\bm{s}^{key\prime} \in \mathbb{R}^{dim}$, where $dim$ is the dimension of a token in Transformers.

\noindent\textbf{Weighted memory-aware aggregation.} For final prediction, we use a score-weighted memory-aware aggregation. To make the predictor aware of the pattern information, we append the corresponding nearest pattern vectors to the claim and key sentence vectors:
\begin{equation}
\label{eq:qsv-concat}
	\bm{v}_i = [\bm{q}^\prime,\bm{s}^{key\prime}_i(q,d),\bm{m}_{j}]
\end{equation}
where $i\!=\!1,\mydots,k_2$. $j\!=\!\mathop{\arg\min}_{k} \left\|\bm{m}_k\!-\!\bm{r}_{s^{key}_i,q}\right\|_2$.

Intuitively, a sentence with higher score should be attended more. Thus, the concatenated vectors (Eq.~\eqref{eq:qsv-concat}) are weighted by the relevance scores from Eq.~\eqref{eq:score_all} (normalized across the top-$k_2$ sentences). The weighted aggregating vector is fed into a MLP which outputs the probability that $d$ fact-checks $q$:
\begin{equation}
	scr^\prime(q,s_i^{key}) = \mathrm{Normalize}\big(scr(q,s_i^{key})\big)
\end{equation}
\begin{equation}\label{eq:final-prediction}
	\hat{y}_{q,d}=\mathrm{MLP}\Big(\sum_{i=1}^{k_2}scr^\prime(q,s_i^{key}) \bm{v}_i\Big)
\end{equation}
where $\hat{y}_{q,d} \in [0,1]$. If $\hat{y}_{q,d}>0.5$, the model predicts that $d$ fact-checks $q$, otherwise does not. The loss function is cross entropy:
\begin{equation}\label{eq:celoss}
	\mathcal{L}_{M}=\mathrm{CrossEntropy}(\hat{y}_{q,d},y_{q,d})
\end{equation}
where $y_{q,d} \in \{0,1\}$ is the ground truth label. $y_{q,d}=1$ if $d$ fact-checks $q$ and $0$ otherwise. The predicted values are used to rank all $k_1$ candidate articles retrieved in the first stage.

\subsection{Training \model}
We summarize the training procedure of \model~in Algorithm~\ref{algo}, including the pretraining of ROT, the initialization of PMB, the training of ARP, and the epoch-wise update of PMB.

\begin{algorithm}[ht] 
\caption{\model~Training Procedure} 
\label{algo}
\begin{algorithmic}[1] 
\Require 
Training set {$\mathcal{T}=[(q_0,d_{00}),\mydots,(q_0,d_{0k_1}),$ $\mydots,(q_n,d_{nk_1})]$ where the $k_1$ candidate articles for each claim are retrieved by BM25.}
\State Pre-train \texttt{ROUGE}-guided Transformer.
\State Initialize the Pattern Memory Bank (PMB).
\For {each epoch}
	\For {$(q,d)$ in $\mathcal{T}$}
		\State {\color{gray} {//} Key Sentence Identification}
		\State Calculate $scr_Q(q,s)$ via ROT and $scr_P(q,s)$ via PMB.
		\State Calculate $scr(q,s)$ using Eq.~\eqref{eq:score_all}.
		\State Select key sentences $\mathcal{K}$.
		\State {\color{gray} {//} Article Relevance Prediction (ARP)}
		\State Calculate $\bm{v}$ for each $s$ in $\mathcal{K}$ and $\hat{y}_{q,d}$.
		\State Update the ARP to minimize $\mathcal{L}_M$.
	\EndFor
	\State Update the PMB using Eq.~\eqref{eq:memory-update}.
\EndFor
\end{algorithmic}
\end{algorithm}

\section{Experiments}\label{sec:exp}
In this section, we mainly answer the following experimental questions:

\noindent\textbf{EQ1:} Can \model~improve the ranking performance of FC-articles given a claim?

\noindent\textbf{EQ2:} How effective are the components of \model, including \texttt{ROUGE}-guided Transformer, Pattern Memory Bank, and weighted memory-aware aggregation in Article Relevance Prediction?

\noindent\textbf{EQ3:} To what extent can \model~identify key sentences in the articles, especially in the longer ones?

\subsection{Data}\label{sec:data}
We conducted the experiments on two real-world datasets. Table~\ref{tab:dataset} shows the statistics of the two datasets. The details are as follows:

\noindent\textbf{Twitter Dataset}

The Twitter\footnote{\url{https://twitter.com}} dataset is originated from \citep{vo2019} and processed by \citet{vo2020}.
The dataset pairs the claims (tweets) with the corresponding FC-articles from Snopes. For tweets with images, it appends the OCR results to the tweets.
We remove the manually normalized claims in Snopes' FC-articles to adapt to more general scenarios. The data split is the same as that in \citep{vo2020}.

\noindent\textbf{Weibo Dataset}

We built the first Chinese dataset for the task of detecting previously fact-checked claims in this article. The claims are collected from Weibo and the FC-articles are from \textit{multiple fact-checking sources} including Jiaozhen, Zhuoyaoji\footnote{\url{https://piyao.sina.cn}}, etc. We recruited annotators to match claims and FC-articles based on basic search results. Appendix~\ref{apd:dataset} introduce the details.

\begin{table}[t]
\caption{Statistics of the Twitter and the Weibo dataset. \#: Number of. C-A Pairs: Claim-article pairs.}
\label{tab:dataset}
{\small
\setlength\tabcolsep{3.2pt}
\begin{tabular}{l|rrr|rrr}
\hline
\multirow{2}{*}{\textbf{Dataset}} & \multicolumn{3}{c|}{\textbf{Twitter}} & \multicolumn{3}{c}{\textbf{Weibo}} \\ \cline{2-7} 
 & Train & Val & Test & Train & Val & Test \\ \hline
\#Claim &  8,002 & 1,000 & 1,001 & 8,356 & 1,192 & 2,386\\ 
\#Articles  & 1,703  & 1,697 & 1,697 & 17,385 & 8,353 & 11,715\\ 
C-A Pairs  &  8,025  & 1,002 & 1,005 & 28,596 & 3,337 &  7,245\\ \hline
\multicolumn{7}{c}{\textbf{Relevant Fact-checking Articles Per Claim}} \\ \hline
Average & 1.003 & 1.002 & 1.004 & 3.422 & 2.799 & 3.036 \\
Medium & 1 & 1 & 1 & 2 & 1  & 2 \\
Maximum & 2 & 2 & 2 & 50 & 18 & 32 \\ \hline
\end{tabular}
}
\end{table}

\begin{table*}[htbp]
\centering
\caption{Performance of baselines and \model. Best results are in \textbf{boldface}.}
\label{tab:main_exp}
\setlength\tabcolsep{3.5pt}
{\small
\begin{tabular}{l|c|cccccc|cccccc}
\hline
\multirow{3}{*}{\textbf{Method}} & \multirow{3}{*}{\begin{tabular}[c]{@{}c@{}}\textbf{Selecting}\\ \textbf{Sentences?}\end{tabular}} & \multicolumn{6}{c|}{\textbf{Weibo}} & \multicolumn{6}{c}{\textbf{Twitter}} \\ \cline{3-14}
 & & \multirow{2}{*}{MRR} & \multicolumn{3}{c}{MAP@} & \multicolumn{2}{c|}{HIT@} & \multirow{2}{*}{MRR} & \multicolumn{3}{c}{MAP@} & \multicolumn{2}{c}{HIT@} \\ \cline{4-8} \cline{10-14}
 & & & 1 & 3 & 5 & 3 & 5 &  & 1 & 3 & 5 & 3 & 5\\ \hline
BM25 & & 0.709 & 0.355 & 0.496 & 0.546 & 0.741 & 0.760 & 0.522 & 0.460 & 0.489 & 0.568 & 0.527 & 0.568 \\ \hline
BERT &  & 0.834 & 0.492 & 0.649 & 0.693 & 0.850  & 0.863  &  0.895 & 0.875 & 0.890 & 0.890 & 0.908  & 0.909 \\
DuoBERT &  & 0.885 & 0.541 & 0.713 & 0.756 & 0.886 & 0.887 & 0.923 & \textbf{0.921} & 0.922 & 0.922 & 0.923 & 0.923  \\
BERT(Transfer) & \checkmark  & 0.714 & 0.361 & 0.504 & 0.553 & 0.742  & 0.764 &  0.642 & 0.567 & 0.612 & 0.623 & 0.668 & 0.719\\ \hline
Sentence-BERT & \checkmark & 0.750 & 0.404 & 0.543 & 0.589 & 0.810 & 0.861 &  0.794 & 0.701 & 0.775 & 0.785 & 0.864 & 0.905\\ 
RankSVM & \checkmark & 0.809 & 0.408 & 0.607 & 0.661 & 0.887 & 0.917 & 0.846 & 0.778 & 0.832 & 0.840  & 0.898 & 0.930\\ 
CTM  &  &  0.856 &	0.356 &	0.481 &	0.525 & 0.894 &	0.935 & 0.926 &	0.889 &	0.919 &	0.922 & 0.952 & 0.964 \\ \hline
\model & \checkmark & \textbf{0.902} & \textbf{0.542} & \textbf{0.741} & \textbf{0.798} & \textbf{0.934} &	\textbf{0.951} & \textbf{0.931}  & 0.899  & \textbf{0.926}  & \textbf{0.928}  & \textbf{0.957}  & \textbf{0.967} \\ \hline
\end{tabular}
}
\end{table*}

\begin{table*}[htbp]
\centering
\caption{Ablation study of \model. Best results are in \textbf{boldface}. AG: Ablation Group.}
\label{tab:ablation}
\setlength\tabcolsep{3.5pt}
{\small
\begin{tabular}{cl|cccccc|cccccc}
\hline
 \multirow{3}{*}{\textbf{AG}} & \multirow{3}{*}{\textbf{Variant}}  & \multicolumn{6}{c|}{\textbf{Weibo}} & \multicolumn{6}{c}{\textbf{Twitter}} \\ \cline{3-14}
 & &\multirow{2}{*}{MRR} & \multicolumn{3}{c}{MAP@} & \multicolumn{2}{c|}{HIT@} & \multirow{2}{*}{MRR} & \multicolumn{3}{c}{MAP@} & \multicolumn{2}{c}{HIT@} \\ \cline{4-8} \cline{10-14}
 & & & 1 & 3 & 5 & 3 & 5 &  & 1 & 3 & 5 & 3 & 5\\ \hline
- & \model & \textbf{0.902} & \textbf{0.542} & \textbf{0.741} & \textbf{0.798} & 0.934 &	0.951 & \textbf{0.931}  & 0.899  & \textbf{0.926}  & \textbf{0.928}  & \textbf{0.957}  & \textbf{0.967} \\ \hline
\ 1& \textit{w/o \texttt{ROUGE} guidance} & 0.892 &	0.535 &	0.729 &	0.786 &0.925 &	0.943  & 0.929  & \textbf{0.905}  & 0.924  & 0.926 & 0.945  & 0.952 \\ \hline
\ \multirow{3}{*}{2} & \textit{w/ rand mem init}  & 0.879 &	0.516 &	0.700 &	0.753 & 0.912 & 0.935 & 0.897 & 0.860 & 0.890 & 0.893 & 0.922  & 0.938 \\
\ & \textit{w/o mem update} &  0.898 & 0.541 & 0.736 & 0.790 & 0.935 &	0.948 & 0.925 &	0.897 &	0.860 &	0.890 & 0.922 &	0.938  \\
\ & \textit{w/o PMB} & 0.897  & 0.537  & 0.734  & 0.792 & 0.931  & 0.948 &  0.920 & 0.885 & 0.913 & 0.917 & 0.944 &0.960 \\\hline
\  \multirow{2}{*}{3} & \textit{w/ avg. pool} & 0.901 & 	0.540 	&0.739 & 0.796& \textbf{0.938} &	\textbf{0.958} & 0.923&0.892 &0.917 &0.919 & 0.944 &0.954\\ 
\ & \textit{w/o pattern aggr.} & 0.896 &	0.535 & 0.734 &	0.791 & 0.930 &	0.945 & 0.922 & 0.890 & 0.917 & 0.919  & 0.947 & 0.954 \\ \hline
\end{tabular}
}
\end{table*}

\subsection{Baseline Methods}\label{sec:baselines}

\noindent\textbf{BERT-based rankers from general IR tasks}

\textbf{BERT}~\citep{bert}: A method of pretraining language representations with a family of pretrained models, which has been used in general document reranking to predict the relevance.~\citep{reranking-with-bert, birch}

\textbf{DuoBERT}~\citep{duobert}: A popular BERT-based reranker for multi-stage document ranking. Its input is a query and a pair of documents. The pairwise scores are aggregated for final document ranking. Our first baseline, BERT (trained with query-article pairs), provides the inputs for DuoBERT.

\textbf{BERT(Transfer)}: As no sentence-level labels are provided in most document retrieval datasets, \citet{simple-bert} finetune BERT with short text matching data and then apply to score the relevance between query and each sentence in documents. The three highest scores are combined with BM25 score for document-level prediction.

\noindent\textbf{Rankers from related works of our task}

\textbf{Sentence-BERT}: \citet{shaar2020} use pretrained Sentence-BERT models to calculate cosine similarity between each sentence and the given claim. Then the top similarity scores are fed into a neural network to predict document relevance.

\textbf{RankSVM}: A pairwise RankSVM model for reranking using the scores from BM25 and sentence-BERT (mentioned above), which achieves the best results in \citep{shaar2020}.

\textbf{CTM}~\citep{vo2020}: This method leverages GloVe and ELMo to jointly represent the claims and the FC-articles for predicting the relevance scores. Its multi-modal version is not included as \model~focuses on key textual information.

\subsection{Experimental Setup}
\noindent\textbf{Evaluation Metrics.} As this is a binary retrieval task, we follow~\citet{shaar2020} and report Mean Reciprocal Rank (MRR), Mean Average Precision@$k$ (MAP@$k$, $k=1,3,5$) and HIT@$k$ ($k=3,5$). See equations in Appendix~\ref{apd:eval-metrics}.

\noindent\textbf{Implementation Details.} In \model, the ROT and ARP components have one and eleven Transformer layers, respectively. The initial parameters are obtained from pretrained BERT models\footnote{We use \texttt{bert-base-chinese} for Weibo and \texttt{bert-base-uncased} for Twitter.}. Other parameters are randomly initialized. The dimension of claim and sentence representation in ARP and pattern vectors are $768$. Number of Clusters in PMB $K$ is $20$. Following~\citep{shaar2020} and~\cite{vo2020}, we use $k_1=50$ candidates retrieved by BM25. $k_2=3$ (Weibo, hereafter, W) / $5$ (Twitter, hereafter, T) key sentences are selected. We use \texttt{Adam}~\citep{adam} for optimization with $\epsilon=10^{-6}, \beta_1=0.9, \beta_2=0.999$. The learning rates are $5\times 10^{-6}$ (W) and $1\times 10^{-4}$ (T). The batch size is $512$ for pretraining ROT, $64$ for the main task.  According to the quantiles on training sets, we set $t_{low}=0.252$ (W) / $0.190$ (T), $t_{high}=0.295$ (W) / $0.227$ (T). The following hyperparameters are selected according to the best validation performance: $\lambda_R=0.01$ (W) / $0.05$ (T), $\lambda_Q=0.6$, $\lambda_P=0.4$, and $\lambda_m=0.3$.
The maximum epoch is $5$. All experiments were conducted on NVIDIA V100 GPUs with PyTorch~\citep{pytorch}.
The implementation details of baselines are in Appendix \ref{apd:imp-baselines}.

\subsection{Performance Comparison}
To answer \textbf{EQ1}, we compared the performance of baselines and our method on the two datasets, as shown in Table \ref{tab:main_exp}. We see that: (1) \model~ourperforms all compared methods on the two datasets (the exception is only the MAP@1 on Twitter), which indicates that it can effectively find related FC-articles and provide evidence for determining if a claim is previously fact-checked. (2) For all methods, the performance on Weibo is worse than that on Twitter because the Weibo dataset contains more claim-sentence pairs (from multiple sources) than Twitter and is more challenging. Despite this, \model's improvement is significant. (3) BERT(Transfer), Sentence-BERT and RankSVM use transferred sentence-level knowledge from other pretext tasks but did not outperform the document-level BERT. This is because FC-articles have their own characteristics, which may not be covered by transferred knowledge. In contrast, our observed characteristics help \model~achieve good performance. Moreover, \model~is also efficiency compared to BERT(Transfer), which also uses 12-layer BERT and selects sentences, because our model uses only one layer for all sentences (other 11 layers are for key sentences), while all sentences are fed into the 12 layers in BERT(Transfer).

\subsection{Ablation Study}
To answer \textbf{EQ2}, we evaluated three ablation groups of \model's variants (AG1$\sim$AG3) to investigate the effectiveness of the model design.\footnote{We do not run MTM without sentence selection due to its high computational overhead which makes it unfeasible for training and inference.} Table~\ref{tab:ablation} shows the performance of variants and \model.

\textbf{AG1: With vs. Without \texttt{ROUGE}.}
 The variant removes the guidance of \texttt{ROUGE} (\model~\textit{w/o \texttt{ROUGE} guidance}) to check the effectiveness of \texttt{ROUGE}-guided finetuning. The variant performs worse on Weibo, but MAP@1 slightly increases on Twitter. This is probably because there are more lexical overlapping between claims and FC-articles in the Weibo dataset, while most of the FC-articles in the Twitter dataset choose to summarize the claims to fact-check. 

\textbf{AG2: Cluster-based Initialization vs. Random Initialization vs. Without update vs. Without PMB.}
The first variant (\model~\textit{ w/ rand mem init}) uses random initialization and the second (\model~\textit{ w/o mem update}) uses pattern vectors without updating. The last one (\model~\textit{ w/o PMB}) removes the PMB. We see that the variants all perform worse than \model~on MRR, of which \textit{w/ rand mem init} performs the worst. This indicates that cluster-based initialization provides a good start and facilitates the following updates while the random one may harm further learning.
 
\textbf{AG3: Score-weighted Pooling vs. Average pooling, and With vs. Without pattern vector.} The first variant, \model~\textit{w/ avg. pool}, replace the score-weighted pooling with average pooling. The comparison in terms of MRR and MAP shows the effectiveness of using relevance scores as weights.
The second, \model~\textit{w/o pattern aggr.}, does not append the pattern vector to claim and sentence vectors before aggregation. It yields worse results, indicating the patterns should be taken into consideration for final prediction.

\subsection{Visualization of Memorized Patterns}
\begin{figure}[t]
	\centering
	\includegraphics[width=0.92\linewidth]{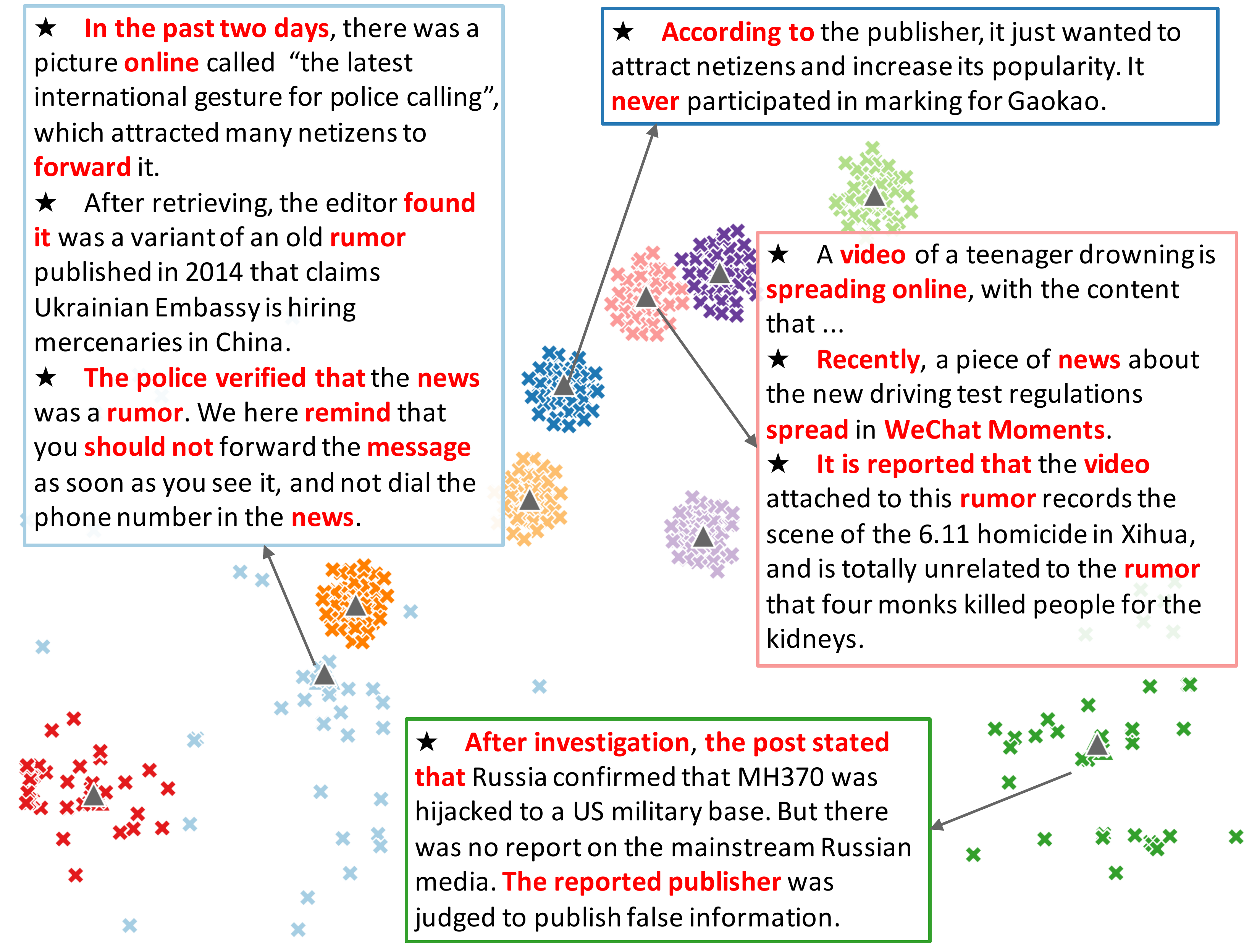}
	\caption{Visualization of pattern vectors ($\blacktriangle$) and near residual embeddings (\ding{54}). The sentences are translated from Chinese.}
	\label{fig:visualization}
\end{figure}

To probe what the PMB summarizes and memorizes, we selected and analyzed the key sentences corresponding to the residual embeddings around pattern vectors. Figure~\ref{fig:visualization} shows example sentences where highly frequent words are in boldface. These examples indicate that the pattern vectors do cluster key sentences with common patterns like ``...spread in WeChat Moments''.

\begin{figure}[tbp]
\centering
\includegraphics[width=0.92\linewidth]{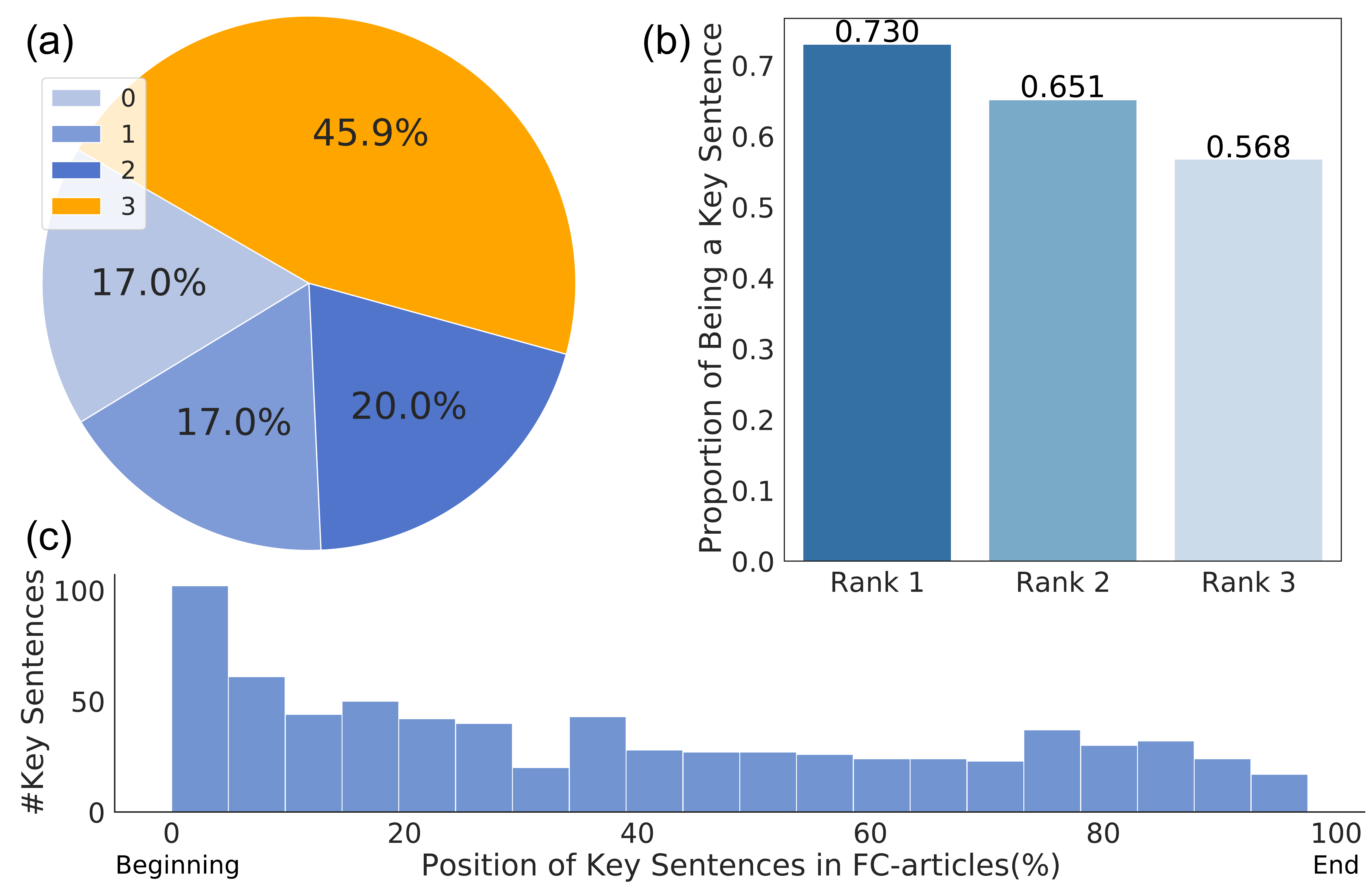}
\caption{Results of human evaluation. (a)The proportion of the FC-articles where \model~found $\{0,1,2,3\}$ key sentences. (b) The proportion of key sentences at rank $\{1,2,3\}$. (c) The positional distribution of key sentences in the FC-articles.}
\label{fig:he}
\end{figure}

\subsection{Human Evaluation and Case Study}
The quality of selected sentences cannot be automatically evaluated due to the lack of sentence-level labels. To answer \textbf{EQ3}, we conducted a human evaluation. We randomly sampled 370 claim-article pairs whose articles were with over 20 sentences from the Weibo dataset. Then we showed each claim and top three sentences selected from the corresponding FC-article by \model. Three annotators were asked to check if an auto-selected sentence helped match the given query and the source article (i.e., key sentences). Figure~\ref{fig:he} shows (a) \model~hit at least one key sentence in 83.0\% of the articles; (b) 73.0\% of the sentences at Rank 1 are key sentences, followed by 65.1\% at Rank 2 and 56.8\% at Rank 3. This proves that \model~can find the key sentences in long FC-articles and provide helpful explanations. We also show the positional distribution in Figure~\ref{fig:he}(c), where key sentences are scattered throughout the articles. Using \model~to find key sentences can save fact-checkers' time to scan these long articles for determining whether the given claim was fact-checked.

\begin{figure}[t]
	\centering
	\includegraphics[width=\linewidth]{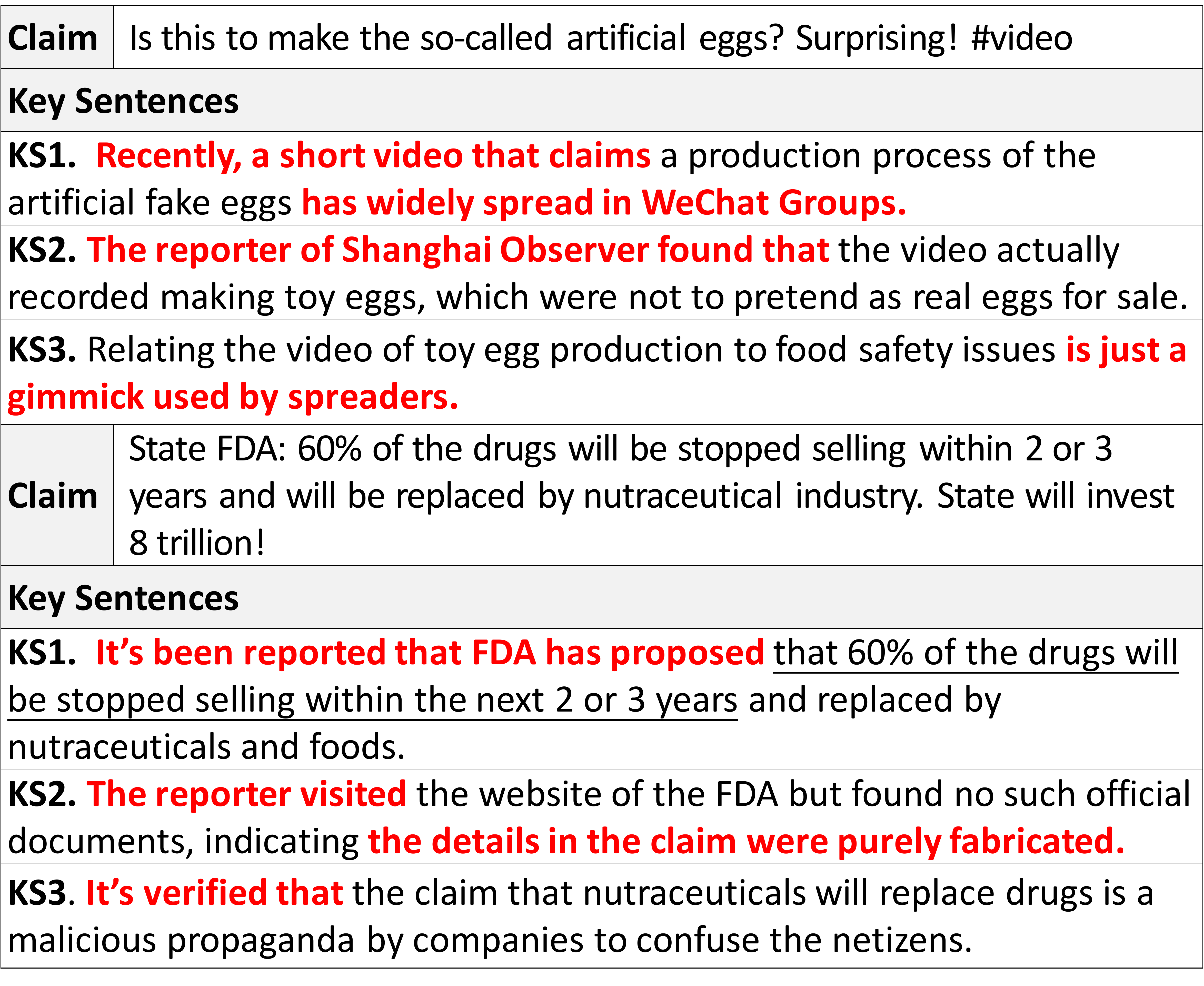}
	\caption{Cases in the set of human evaluation. Quotations are underlined and patterns are in {\color{red} \textbf{boldface}}.}
	\label{fig:cases}
\end{figure}

Additionally, we exhibit two cases in the evaluation set in Figure~\ref{fig:cases}. These cases prove that \model~found the key sentences that correspond to the characteristics described in Section~\ref{sec:intro}. Please refer to~Appendix \ref{apd:error} for further case analysis.

\section{Conclusions}

We propose \model~to select from fact-checking articles key sentences that introduce or debunk claims. These auto-selected sentences are exploited in an end-to-end network for estimating the relevance of the fact-checking articles w.r.t. a given claim. Experiments on the Twitter dataset and the Weibo dataset (newly constructed) show that \model ~outperforms the state of the art. Moreover, human evaluation and case studies demonstrate that the selected sentences provide helpful explanations of the results. 

\section*{Acknowledgments}
The authors thank Guang Yang, Tianyun Yang, Peng Qi and anonymous reviewers for their insightful comments. Also, we thank Rundong Li, Qiong Nan, and other annotators for their efforts. This work was supported by the National Key Research and Development Program of China (2017YFC0820604), the National Natural Science Foundation of China (U1703261), and the Fundamental Research Funds for the Central Universities and the Research Funds of Renmin University of China (No. 18XNLG19). The corresponding authors are Juan Cao and Xirong Li.

\section*{Broader Impact Statement}
Our work involves two scenarios that need the ability to detect previously fact-checked claims: (1) For social media platforms, our method can check whether a newly published post contains false claims that have been debunked. The platform may help the users to be aware of the text's veracity by providing the key sentences selected from fact-checking articles and their links. (2) For manual or automatic fact-checking systems, it can be a filter to avoid redundant fact-checking work. When functioning well, it can assist platforms, users, and fact-checkers to maintain more credible cyberspace. But in the failure cases, some well-disguised claims may escape. This method functions with reliance on the used fact-checking article databases. Thus, authority and credibility need to be carefully considered in practice. We did our best to make the new Weibo dataset for academic purpose reliable. Appendix~\ref{apd:dataset} introduces more details.

\bibliographystyle{acl_natbib}
\bibliography{acl2021}

\appendix
\section{Constructing the New Weibo Dataset}
\label{apd:dataset}
To construct datasets for fact-checked claim detection on social media, we need to (1) collect the fact-checked claims (social media posts); (2) collect fact-checking articles (FC-articles); and (3) generate claim-article pairs. 
\paragraph{Collection.} In Step (1), we used posts whose labels are \emph{fake} from the datasets for fake news detection~\citep{dual-emotion, newsverify}, because their labels were determined by fact-checking. In Step (2), we crawled fact-checking articles from multiple sources to enrich the article base. The sources are partially listed in Table~\ref{tab:sources} due to the space limit. For the claims and articles which contained much text in the attached images, we recognized the text using OCR service on Baidu AI platform\footnote{\url{https://ai.baidu.com/tech/ocr}}. Note that we only crawled the claims and articles that were publicly available at the crawling time. To protect privacy, the publishers' names were removed. However, we preserved names and offensive words in the main text because they were crucial for summarizing the events and performing the matching process.
\paragraph{Annotation.} In Step (3), we performed a model-assisted human annotation. We first duplicated the data collected in Step (1) and (2) and then used BM25 to retrieve the relevant FC-articles as candidates with the claims as queries. Twenty-six annotators (postgraduates) were instructed (by a Chinese guideline with examples written by the first author) to check whether the candidates did fact-check the given claims. We dropped the claims that are annotated as irrelevant to all candidates. For claims that were with highly overlapping candidates but different annotation results, the authors manually checked and corrected the wrongly annotated samples. 

\begin{table*}[ht]
\caption{Part of the Sources of fact-checking articles in the Weibo dataset.}
\label{tab:sources}

{\small
\begin{tabular}{p{0.18\linewidth}p{0.4\linewidth}p{0.33\linewidth}}
\hline
Source & Description & URL \\ \hline
Jiaozhen & A fact-checking platform operated by Tencent. & \url{https://fact.qq.com/}, \url{https://new.qq.com/omn/author/5107513} \\
Liuyanbaike & A debunking website operated by Guokr. & \url{http://www.liuyanbaike.com/} \\
Baidu Piyao & A fact-checking account operated by Baidu. & \url{https://author.baidu.com/home?app_id=15060} \\
ScienceFacts & A platform to fact-check scientific claims supported by China Association for Science and Technology & \url{https://piyao.kepuchina.cn/} \\
Qiuzhen & A fact-checking column of People's Daily Online & \url{http://society.people.com.cn/GB/229589/index.html} \\
Dingxiang Doctor & A platform for doctors and experts in life science & \url{https://dxy.com/} \\
China Joint Internet Rumor-Busting Platform & A platform operated by Cyberspace Administration of China & \url{http://www.piyao.org.cn/} \\
Zhuoyaoji & Sina News official fact-checking account & \url{http://piyao.sina.cn/}, \url{https://weibo.com/u/6590980486} \\ 
Weibo Piyao & Weibo official fact-checking account & \url{https://weibo.com/weibopiyao} \\ \hline
\end{tabular}
}
\end{table*}

\section{Calculation of Evaluation Metrics}
\label{apd:eval-metrics}
Assume that query set $Q$ has $|Q|$ queries and the $i^{th}$ query has $n_i$ relevant documents. We calculate the evaluation metrics using the following equations:
\begin{equation}
	\mathrm{MRR}=\frac{1}{|Q|}\sum_{i=1}^{|Q|}\frac{1}{rank_i}
\end{equation}
where $rank_i$ refers to the rank position of the first relevant answer for the $i^{th}$ query in the corresponding retrieving result.~\citep{mrr}
\begin{equation}
	\mathrm{MAP}@k=\frac{1}{|Q|}\sum_{i=1}^{|Q|}\frac{1}{n_i}\sum_{j=1}^{n_i} P_i(j)rel_i(j)
\end{equation}
where $P_i(j)$ is the proportion of returned documents in the top-$j$ set for the $i^{th}$ query that are relevant. $rel_i(j)$ is an indicator function equaling $1$ if the document at rank $j$ in the returned list for the $i^{th}$ query is relevant and $0$ otherwise.~\citep{map-new}
\begin{equation}
	\mathrm{HIT}@k=\frac{1}{|Q|}\sum_{i=1}^{|Q|}has_i(k)
\end{equation}
where $has_i(k)$ is an indicator function equaling $1$ if $rank_i\leq k$ and $0$ otherwise.~\citep{hit}

Note that we guarantee that a query has at least one relevant document in its candidate list, so the corner case of empty ground truth set is ignored. 

\section{Implementation of BM25 and Baselines}
\label{apd:imp-baselines}

\paragraph{BM25:} The articles were indexed with \texttt{gensim}~\citep{gensim}.
\paragraph{BERT:} We finetuned the last Transformer layer of \texttt{bert-base-chinese} for Chinese and \texttt{bert-base-uncased} for English. Following the commonly used strategy (e.g., \citealp{bert-trunc}), we truncated the sequences to the maximum length of $512$. The maximum length of claims is the same as \model~and the rest tokens are from articles.
\paragraph{DuoBERT}: We used top 20 articles from the results of BERT as candidates to construct article pairs. For each article, the score is obtained by summing its pairwise scores. The used pretrained models are the same as BERT (mentioned above) and we finetuned the layers except the embedding layer and the first Transformer layer.
\paragraph{BERT(Transfer):} For the Twitter data, we used the models provided in Birch~\citep{birch} that was finetuned on TREC Microblog Track data~\citep{trec-microblog}; for the Weibo data, we used LCQMC dataset~\citep{LCQMC} containing 260,068 text pairs to finetune \texttt{bert-based-chinese} for 20 epochs. Considering the value difference between BM25 and BERT scores, the weight of BM25 score was learned by grid search in $[0,1]$ but the weights of others were in $[0,5]$. The step size was $0.1$. We got the best results with BM25 weight $= 0.2$ (Weibo) / $0.1$ (Twitter) and the weights of top-3 sentences $=1.2, 0.4, 0.9$ (Weibo) / $4.8, 4, 2.5$ (Twitter), respectively.
\paragraph{Sentence-BERT:}  We used the base versions in \texttt{Sentence-Transformers}~\citep{sentence-bert}
 to obtain the embeddings against the claims and sentences. Specifically, we used \texttt{stsb-xlm-r-multilingual}~\citep{sbert-multilingual} for the Weibo data and \texttt{stsb-bert-base} for Twitter\footnote{\url{https://www.sbert.net/docs/pretrained_models.html}} . According to \citet{shaar2020}, we calculated the cosine similarity of each claim-sentence pair and fed the top-$5$ scores into a simple neural network (20-ReLU-10-ReLU) for classification. We trained the model for $20$ epochs with class weighted cross entropy as the loss function. The class weights were calculated across the dataset~\citep{class-weight}.
\paragraph{RankSVM:} We combined the scores and their reciprocal ranks obtained from Sentence-BERT models and BM25. Then we fed them into a RankSVM\footnote{\url{http://www.cs.cornell.edu/people/tj/svm_light/svm_rank.html}}~\citep{ranksvm} for classification. We used Sentence-BERT models trained with $\{3,4,5,6\}$ sentences for Twitter and those trained with $\{6,7,8,9\}$ sentences for Weibo. We kept the default settings in the package.
\paragraph{CTM:} For the Twitter dataset, we followed~\citep{vo2020} to use \texttt{glove.6B}\footnote{\url{https://nlp.stanford.edu/projects/glove/}}~\citep{glove} and the \texttt{ELMo Original (5.5B)}\footnote{\url{https://allennlp.org/elmo}}~\citep{elmo}; for the Weibo data, we used \texttt{sgns.weibo.bigram-char}\footnote{\url{https://github.com/Embedding/Chinese-Word-Vectors}}~\citep{chinese-embedding} and \texttt{simplified-Chinese ELMo}\footnote{\url{https://github.com/HIT-SCIR/ELMoForManyLangs}}~\citep{elmo-chinese, elmo-chinese2}. We kept the default settings provided by the authors\footnote{\url{https://github.com/nguyenvo09/EMNLP2020}}.

\begin{figure}[t]
	\centering
	\includegraphics[width=0.46\textwidth]{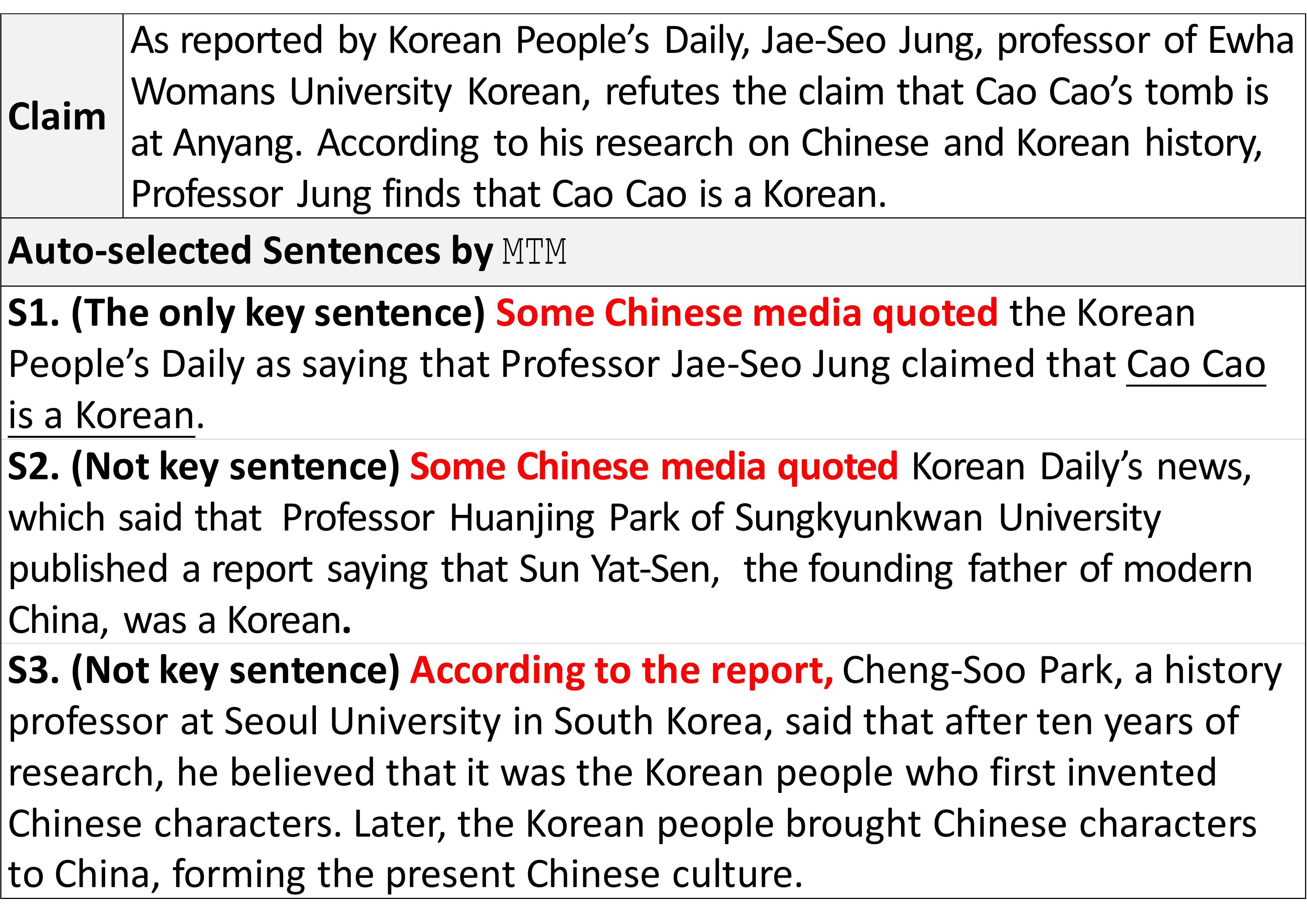}
	\caption{A case with only one key sentence being hit by \model. Patterns are in {\color{red} \textbf{boldface}}.}
	\label{fig:further_case1}
\end{figure}

\begin{figure}[t]
	\centering
	\includegraphics[width=0.46\textwidth]{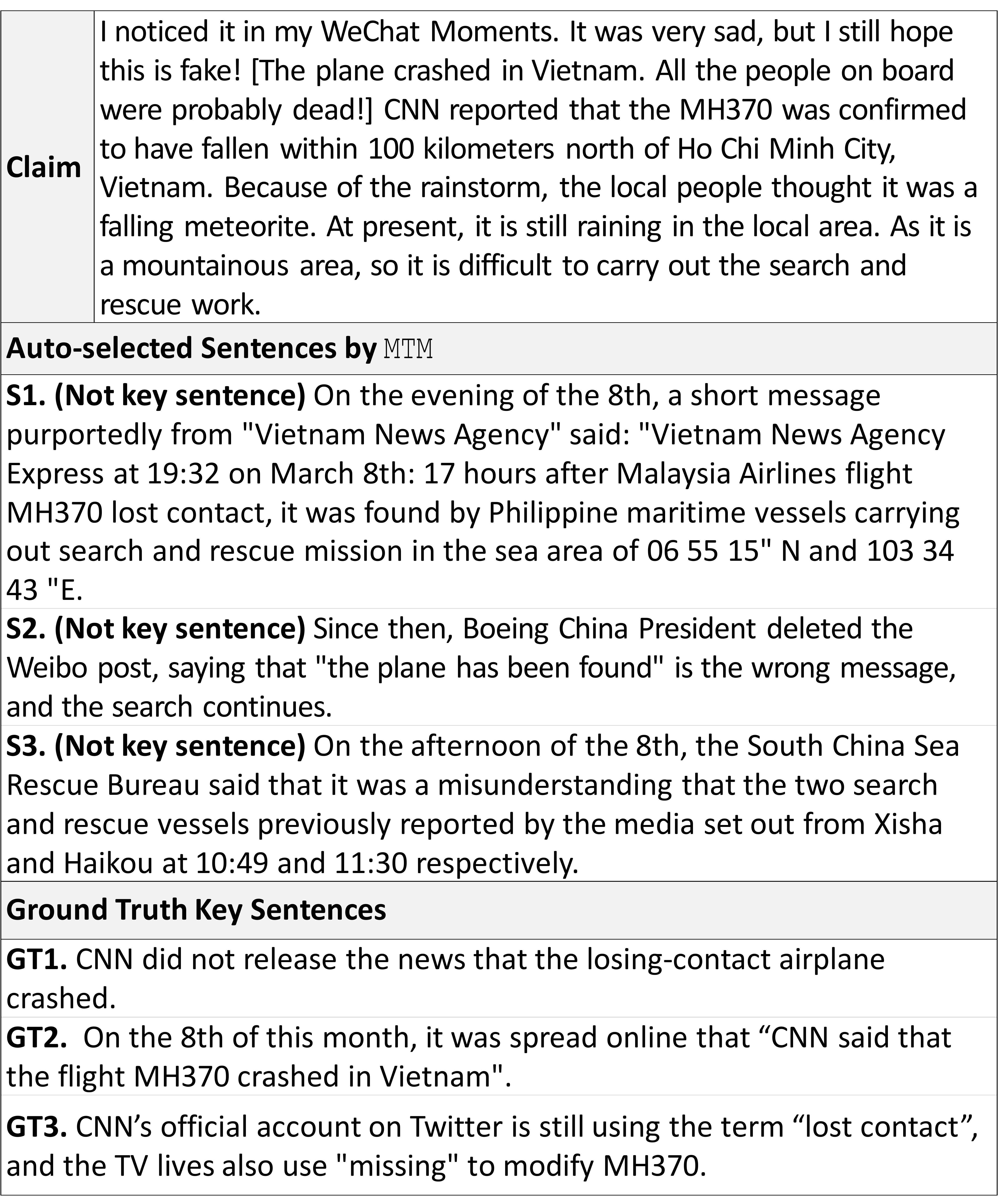}
	\caption{A case with no key sentence being hit by \model.}
	\label{fig:further_case2}
\end{figure}

\section{Further Case Analysis}
\label{apd:error}
We reviewed the fact-checking articles in the set for human evaluation wherein \model~hit less than two key sentences. We here exhibit two situations that make \model~ did not perform well: (1) In Figure~\ref{fig:further_case1}, the claim is about where Cao Cao was born. \model~found three sentences with significant patterns (shown in boldface). However, only S1 is related to the claim. S2 and S3 introduce similar but irrelevant claims. This is because that the fact-checking article is actually a collection of rumors about South Korea on the Chinese social media. The claims in this article are all similar to each other, and thus, to differentiate them needs more delicate semantic understanding. (2) Figure~\ref{fig:further_case2} shows a case where \model~found no key sentence from the article. We append the key sentences selected manually below. We speculate that the failure is due to the length of the given claim. The claim is longer than general posts on Weibo and contains many details, making the model lose focus on the key elements of the event description. Thus, S1 describing another news about MH370's activity in Vietnam was selected, instead of the ground truth sentences. To achieve better performance, future work may consider improving the semantic modeling and summarizing key information from both fact-checking articles and claims.

\end{document}